# ActiNet: An Open-Source Tool for Activity Intensity Classification of Wrist-Worn Accelerometry Using Self-Supervised Deep Learning

Aidan Acquah, Shing Chan, Aiden Doherty

## Abstract

The use of accurate and reliable open-source human activity recognition (HAR) models on passively collected wrist-accelerometer data is essential in large-scale epidemiological studies that investigate the association between physical activity and health outcomes. While self-supervised learning has generated considerable excitement in improving HAR, the extent to which these models, coupled with hidden Markov models (HMMs), would make a tangible improvement to classification performance and the effect this may have on the predicted daily activity intensity compositions is unknown. Using up to 24 hours of wrist-worn accelerometer data from 151 CAPTURE-24 participants (aged 18 – 91, mean age 42, 66% female), we trained the ActiNet model, consisting of HARNet, a self-supervised, 18-layer, modified ResNet-V2 model, followed by hidden Markov model (HMM) smoothing, to classify labels of activity intensity. The performance of this model, evaluated using 5-fold stratified group cross-validation, was then compared to a baseline random forest (RF) + HMM, established in existing literature. Differences in performance and classification outputs were compared with different subgroups of age and sex within the CAPTURE-24 population. The ActiNet model was able to distinguish labels of activity intensity with a mean macro F1 score of 0.80 and a mean Cohen's kappa score of 0.85. This exceeded the performance of the RF + HMM, trained and validated on the same dataset, with mean scores of 0.76 and 0.80, respectively. The improvements in performance were consistent across subgroups of age and sex. These findings encourage the use of ActiNet for the extraction of activity intensity labels from wrist-accelerometer data in future epidemiological studies.

# Introduction

The breakdown of the proportion of the day spent in different activity intensities; sleep, sedentary, light and moderate-vigorous activity, can give useful insights into health outcomes, as seen in various studies[1–5]. Improved estimation of these daily activity intensity compositions can better inform public health messaging, such as that of the World Health Organisation (WHO)[4,6,7]. Traditionally, these activity levels were measured using activity diaries, however, this was prone to self-recall bias, encouraging the adoption of wearable accelerometers for more objective activity monitoring[8,9]. One popular approach to extract meaningful labels of activity intensity from accelerometer data is the use of open-source, human activity recognition (HAR) machine learning (ML) tools, such as the pip package accelerometer[10], consisting of a random forest (RF)[11], with hidden Markov model (HMM)[12] smoothing, used in various epidemiological studies[1–3]. The open-source nature of such tools enable widespread use and collaborative development, with a transparent codebase, that supports verifiable and reproducible research outputs[13]. Self-supervised learning (SSL) appears to provide a transformative step forward in ML HAR[14–17], with numerous research studies showing that a HARNet-based model, consisting of an SSL pre-trained, 18-layer, modified ResNet-V2[18] model can improve classification accuracy over other ML approaches[14,19,20].

However, it remained uncertain how the HARNet model, when combined with HMM smoothing, would perform in the classification of activity intensity, as compared to an existing open-source ML model in wide circulation. For the new model created, it was also unknown whether it may introduce some biases in classification performance, particularly in the classification of certain behaviours, or within subgroups of age or sex, influencing the generalisability of the model to the wider population. Furthermore, it was unknown how the difference in classification performance may influence the activity intensity composition classified for each participant; how this may differ from the baseline model, and whether this would substantially reduce the error to the ground-truth labels.

We therefore developed the open-source tool, ActiNet[21], a HARNet-based model, followed by HMM smoothing, trained to classify labels of activity intensity from wrist-worn accelerometer data. We evaluated the performance of the model by comparing classification outputs to ground-truth labels, and compared this performance to a baseline RF + HMM model, modelled after the existing accelerometer pip package. We also observed how the performance of these models differed within subgroups of age group and sex within the study population. Finally, we compared the volume of classification outputs for each activity intensity label between the ground-truth labels, baseline and ActiNet models.

# Related Work

## Activity Intensity Labels

The activity recognition models explored in this study map raw wrist-worn accelerometry into interpretable labels of activity intensity. Activity intensity was defined in this study using the Walmsley mapping[1], which converts the full spectrum of human behaviours into four categories: sleep, sedentary behaviour, light activity, and moderate-vigorous activity. Where applicable, these categories align with conventional metabolic equivalent task (MET) thresholds, and are defined as follows:

- Sleep: non-waking behaviour
- Sedentary behaviour: waking behaviour, spent sitting, lying or in a reclined posture[22]
- Light activity: waking, non-sedentary behaviour at $< 3$ METs
- Moderate-vigorous activity: waking, non-sedentary behaviour at $\geq 3$ METs[23]

## Open-source activity intensity classification models

This study focused on the development of an open-source model for classifying activity intensity from wrist-worn accelerometry. Early approaches to human activity recognition relied predominantly on cutpoint-based methods, in which summary acceleration metrics, such as Euclidean Norm Minus One (ENMO)[24], were compared against predefined threshold values to assign activity intensity labels.

This methodology was popularised by the GGIR package[25], which by default calculated ENMO for 5-second windows of accelerometry, using a threshold of 40mg to distinguish sedentary behaviour from light activity, and 100mg to distinguish light from moderate-vigorous activity[25]. More sophisticated algorithms have also been incorporated to detect sleep periods[26,27].

More recent works have adopted ML approaches, particularly random forest-based classifiers, to improve classification performance and model generalisability[28]. Examples include the OxWearables random forest model, available in the pip package "accelerometer"[10], the Actimetric package Trost model[29], and the GGIR extension Sundararajan sleep model[30]. These models leverage engineered time- and frequency-domain features, that offer a more complete picture of wearable behaviour than just exceeding thresholds in average acceleration, and have demonstrated improved performance relative to simpler cutpoint-based methods/heuristics[1,30].

The CAPTURE-24 dataset[31] is currently the largest openly available dataset of annotated, free-living, wrist-worn accelerometry, and has therefore enabled the training and validation of activity intensity classifiers. Annotations for this dataset were derived by participants wearing an OMG Life Autographer camera, which captured images every 20-40 seconds[32]. These images were subsequently used to infer detailed behavioural labels, which can then be further simplified to labels of activity intensity. Using this dataset, Chan et al. (2024)[32] benchmarked various ML models, trained to classify activity from wrist accelerometer data, using a range of approaches to extract signal features, classify signal patterns, and account for temporal dependencies. This evaluation demonstrated both the advantages of hidden Markov model (HMM) smoothing, and the potential for deep learning models, particularly a convolutional neural network (CNN) combined with an HMM, to outperform other ML approaches, such as random forest-based models, and temporal deep learning models like long short term memory (LSTM)/recurrent neural network (RNN)-based models[32].

A promising, open-source, deep learning approach for improving HAR performance is the HARNet model, introduced by Yuan et al. (2024)[14]. HARNet uses a modified 18-layer ResNet-V2[18] architecture, and was trained on the UK Biobank physical activity monitoring dataset[33], consisting of roughly 700,000 person-days of wrist-worn accelerometry data. The model was trained using self-supervised learning, designed to recognise signal modifications, including flipping the arrow of time, enabling the model to learn informative signal representations. The learned representations were shown to separate activity intensity classes better than the hand-crafted features typically used in random forest models. This suggests that HARNet-based architectures may provide improved representation learning for activity intensity classification compared with existing open-source machine-learning approaches.

Recent studies have investigated the use of deep learning approaches for HAR in free-living accelerometry. Sameh et al. (2025)[34] compared the use of traditional ML models based on hand-crafted features with deep learning techniques to extract 24-hour movement behaviours from wrist-worn accelerometry in the CAPTURE-24 dataset. Their findings indicate that while deep learned features offer moderate improvements, traditional models such as random forests can still achieve competitive, and in some cases, superior, classification performance in this setting. Notably, these deep learning models did not use self-supervised learning. Logacjov et al. (2024)[35] introduced SelfPAB, an open-source, self-supervised, spectrogram-based transformer model for HAR. This approach demonstrated substantial improvements over an XGBoost baseline for classifying activities of daily living using thigh and lower back accelerometers, highlighting the potential of leveraging unlabelled data, similarly seen in Yuan et al. (2024)[14]. A recent review by Zi el al. (2026)[36], collected evidence from 40 studies, further supporting the growing role of deep learning in enhancing activity intensity classification, particularly in free-living environments. Collectively, these findings suggest a shift in the field towards representation learning approaches, where performance gains are increasingly driven by large-scale data and self-supervised learning rather than purely supervised model design.

# Methods

## Datasets

This study was performed on the CAPTURE-24 dataset[31], consisting of wrist-worn accelerometer data collected from 151 participants from the Oxfordshire area[32]. These participants were aged 18 – 91 years, with a mean age of 42 years, and 66% were female. The accelerometer data was collected using an AX3 activity device, at a sampling frequency of 100Hz for roughly one day. The ground-truth annotations for activities performed by participants during wear were obtained from the combination of sleep diaries, to track sleep onset and offset, and a chest-mounted OMG Life Autographer camera, capturing images of the participants' surroundings every 20-40 seconds. Annotators have converted these images into labels of activities of daily living, which could then be further simplified to labels of activity intensity using the Walmsley mapping[1]. The CAPTURE-24 dataset is the perfect dataset for this study, as it is the largest collection of publicly available annotated wrist-worn accelerometer data to date[1,32].

## Pre-processing

The raw accelerometer files (.cwa) were read and pre-processed using the actipy pip package[37]. The pre-processing steps ensured the conversion of the accelerometer data into a harmonised, consistent form, usable in this study. First, we detected any periods of non-wear, identified as periods in which the vector magnitude falls below 15m*g* for a period of 90 minutes, for a rolling window of 10 seconds. For all detected non-wear periods, data was removed to avoid erroneous classification of activities. We then auto-calibrated the signal to ensure that the stationary periods of the signal had a magnitude of acceleration equal to 1*g*, following approaches recommended by Van Hees et al. (2014)[38]. Finally, the signal was divided into 30-second windows of behaviour, with ground-truth annotations for each window defined as the most frequent label within the window.

# Models

Within this study, we developed a novel, open-source tool named ActiNet, consisting of an initial classifier block of a HARNet model[14], followed by a hidden Markov model (HMM)[12] for smoothing, and post-hoc sleep correction. We compared this to a baseline model, consisting of a random forest (RF) classifier block, with an HMM and sleep correction, based on the existing accelerometer[10] pip package. These models were both trained on the CAPTURE-24 dataset[31], to classify labels of activity intensity from 30-second windows of wrist-worn accelerometer data. From these output classifications, inferences could be made in comparing the breakdown of activity intensity within different participants/populations.

## HARNet

HARNet is an existing, open-source model, consisting of a modified ResNet-V2[18] with 18 layers of 1D convolutions to serve as a feature extractor, with roughly 10 million parameters, and an output feature vector size of 1,024. This deep-learned feature extractor was trained using self-supervised learning to learn the arrow of time classification task on one day of unlabelled data from roughly 100,000 UK Biobank participants[33], to distinguish the difference between forward and reverse time accelerometer signals. This architecture takes an input vector of 900 x 3, corresponding to a 30Hz, 30-second window, with 3 axes of acceleration[14]. This pre-trained model, made publicly available by Yuan et al. (2024)[14], was downloaded and used in this study.

The final layer of the model, is a fully connected layer with optional SoftMax activation, taking the 1,024 feature inputs, and outputting a classification probability for each possible label for the specific classification task. In this study, we fine-tuned all the weights, both the final layer and the pre-trained deep layers of the model on the CAPTURE-24 dataset[31], to the four possible labels of activity intensity. To do so, we used an Adam optimiser[39] with a learning rate of 0.0001, a random state of 42, a batch size of 1,000, a total number of training epochs of 100, but with an early stopping patience of 5 epochs. We used inverse class weighting to ensure balanced training of all activity intensity labels, with a cross-entropy loss function. During training, three data augmentation strategies were applied: random decimation, performed at scales of 1x, 2x, or 3x; random axis switching; and random axis rotation. The final output classification of the model for the 30-second window was a single label corresponding to the highest value output by the SoftMax layer.

## Random Forest

In order to benchmark the performance of the ActiNet model, we used random forest[11] + HMM models, similar in structure to the widely used accelerometer pip package, to serve as a baseline. From the 100Hz, 30-second pre-processed windows, the model first extracts hand-selected features from the accelerometer signal by the accelerometer pip package[10]. As listed in supplementary table 1, this includes 37 axis-independent features, with signal-related properties such as frequency, correlation and moment, known to have a strong association with different activity intensity levels[1,10].

These extracted features were then fed into a random forest model, with 200 estimators (trees), not minority balanced sampling, a maximum depth for trees of 20, a minimum samples per leaf of $5\times10^{-5}$, and fixed random state of 42.

## Hidden Markov Model

Hidden Markov models (HMMs) were employed in both models to smooth the sequence of classifications derived from consecutive 30-second windows. For each model, the sequence of predicted labels derived from the windowed accelerometer data was treated at the observation sequence. The most probable sequence of latent states underlying these observations was inferred using the Viterbi algorithm[40].

The HMM was parameterised through estimation of the transition and emission probability matrices. The transition matrix represents the probability of transitioning to a given state at the subsequent time step, given the current state. Modifications were required when estimating this matrix using the CAPTURE-24 dataset due to discontinuities in annotation. Specifically, transitions were only included if they corresponded to the expected 30-second interval between adjacent windows. Furthermore, to account for systematic gaps surrounding annotated sleep periods, two additional transitions were introduced for each participant: sleep-to-sedentary and sedentary-to-sleep. This adjustment reduced the concern around sleep-related transitions, due to training the HMM on data with incomplete annotation continuity.

The emission matrix represents the probability of an observed label, as produced by the classification model, given an underlying true state. This was estimated via normalised cross-tabulation of true labels and model outputs on training data. For the random forest, out-of-bag scores were used to obtain unbiased estimates. For the ActiNet model, SoftMax outputs derived from a held-out 20% subset of the training data were used.

Inference of the most likely sequence of latent states was performed using the Viterbi algorithm, given the learned transition and emission matrices. A uniform prior distribution for initial states was assumed. Subsequent state estimates were obtained recursively by maximising the joint probability conditioned on the previous state, current observation, and the learned model parameters.

### Post-Processing

Post-hoc sleep correction was applied to the output sequence of states produced by the HMM. Consistent with the current accelerometer[10] pip package, we required all classified blocks of sleep to last at least one hour. Any sleep classifications that did not lie within such a block were instead converted to sedentary behaviour.

## Training and Evaluation

### Cross Validation

To train and evaluate the models, we used five-fold stratified group cross-validation[41]. Training was stratified by activity intensity label, to ensure roughly even representation of class labels across training and test sets in each fold. Training and test splits were based on groups, that is, participant ID, to ensure that there was no leakage of the same participant's data in both training and testing, giving an accurate representation of the model performance on unseen data.

For the random forest model, the full training set could be used to train the random forest, and out-of-bag scores from the full set, alongside ground-truth labels, were used to train the HMM. For ActiNet, however, another split of training into training and validation sets was required. Nested within each cross-validation fold was a single group shuffle split, used to hold out 20% of participants' data from training the HARNet model. Instead, these participants' data were used for early stopping of training, as well as for producing estimates of classification predictions, used to train the emissions matrix of the HMM. This nested cross-validation approach is illustrated in Supplementary Figure 1.

## Evaluation

*Classification Performance*

The evaluation metrics used within this study, to observe how well the models could correctly identify labels of activity intensity, were accuracy, balanced accuracy, macro F1 and Cohen's kappa score, as well as confusion matrices. To report these performance metrics on the CAPTURE-24 dataset, we aggregated the performance of the model for each participant, for the fold for which they were in the test set and reported the mean ± standard deviation across test participants.

To provide further validity to the model, we evaluated the fully trained and deployed model on external datasets, namely Realworld[42] and WISDM[43]. These datasets contained a smaller amount of wrist-worn accelerometer data, collected in limited environments observing activities of daily living.

*Activity Composition*

The activity composition represents the proportion of time spent in each activity intensity over the period of monitoring. While we were interested in the improved performance of the ActiNet model, it was also critical to confirm that the new model produced accurate estimates of the activity composition, as this is what is extracted for association studies, and public health messaging. We were also interested in understanding the difference in classification output volume of activity labels between the two models, to discover potential changes from adopting this model, over the baseline.

To evaluate the improved identification of activity composition, we compared the total number of hours of each activity label classified by the model, as compared to the ground-truth. We calculated the percentage error for the classification outputs for both models and used boxplots to present the difference in distribution between the two models. We used the mean absolute error (MAE), to compare the two models for each activity intensity label, with paired t-tests to investigate significance.

To further compare differences between the ground-truth activity volumes and predictions made by the baseline and ActiNet models, we used Bland-Altman[44] plots, observing the distribution differences in the proportion of activity labels. For these plots, we observed the mean difference, and upper and lower limits of agreement in the amount of each activity intensity, offering insight into patterned differences in classification outputs by the models. We also calculated the Pearson correlation coefficient[45] to compare the degree of similarity between the model outputs.

*Subgroup Patterns*

For both classification performance and activity composition, we were interested in any biases that may be introduced by the model, with respect to age and/or sex. To investigate this, we compared the output behaviour of both models, within subgroups of age: 18-29, 30-37, 37-52, 53+ and sex: female, male.

All code was run on Python 3.9, with deep learning models run on PyTorch version 1.13.1+cu117.

# Results

The number of 30-second samples taken from 151 participants of the CAPTURE-24 study totalled 318,686, equating to roughly 18 hours of annotated data per person. The annotation labels were unbalanced, with the most frequent label of sedentary behaviour (40% of windows), followed by sleep (36%), light (20%), and moderate-vigorous (5%). While most participants performed a combination of each of these activities throughout the day of monitoring, 16 participants, 11% of the CAPTURE-24 study population, had no data with an annotation label of moderate-vigorous activity.

## Classification Performance

As described in the methods, the ActiNet and baseline RF + HMM models were trained and evaluated using 5-fold cross-validation. The list of participants corresponding to each fold for the cross-validation is in Supplementary Table 2.

The performance metrics for both the baseline and ActiNet models are presented in Table 1. Across these per-participant evaluation metrics, the ActiNet model outperforms the baseline. The ActiNet model produced a mean (± standard deviation) Cohen's kappa score of 0.85 ± 0.11 and macro F1 of 0.80 ± 0.11. In comparison, the baseline model produced a Cohen's kappa of 0.80 ± 0.12, macro F1 of 0.76 ± 0.12. Using a paired t-test, the per-participant improvement in performance metrics was statistically significant, with p values far below 0.001.

*Table 1: Evaluation performance for baseline (random forest + hidden Markov model) and ActiNet models trained on the CAPTURE-24 dataset using nested 5-fold group cross-validation to classify labels of activity intensity. Scores are reported as the mean ± standard deviation for participant scores, with the highest score indicated in bold. P-values extracted from paired t-test comparing the evaluation metrics scores for both models for all participants.*

| Model | Accuracy | Balanced Accuracy | Cohen's kappa | Macro F1 |
|---|---|---|---|---|
| **Baseline** | 0.872 ± 0.080 | 0.792 ± 0.102 | 0.800 ± 0.116 | 0.764 ± 0.116 |
| **ActiNet** | **0.902 ± 0.074** | **0.829 ± 0.097** | **0.847 ± 0.106** | **0.799 ± 0.108** |
| ***p-value*** | *<0.001* | *<0.001* | *<0.001* | *<0.001* |

*Figure 1: Confusion matrices for the performance of CAPTURE-24-trained and evaluated activity recognition models, using nested 5-fold group cross-validation. The baseline model is a random forest-based model, and ActiNet is a HARNet-based model. Both models consist of hidden Markov model smoothing and post hoc sleep correction. Each cell shows the proportion of 30-second windows with a given true activity intensity that were predicted as each activity class, with corresponding window counts in parentheses.*

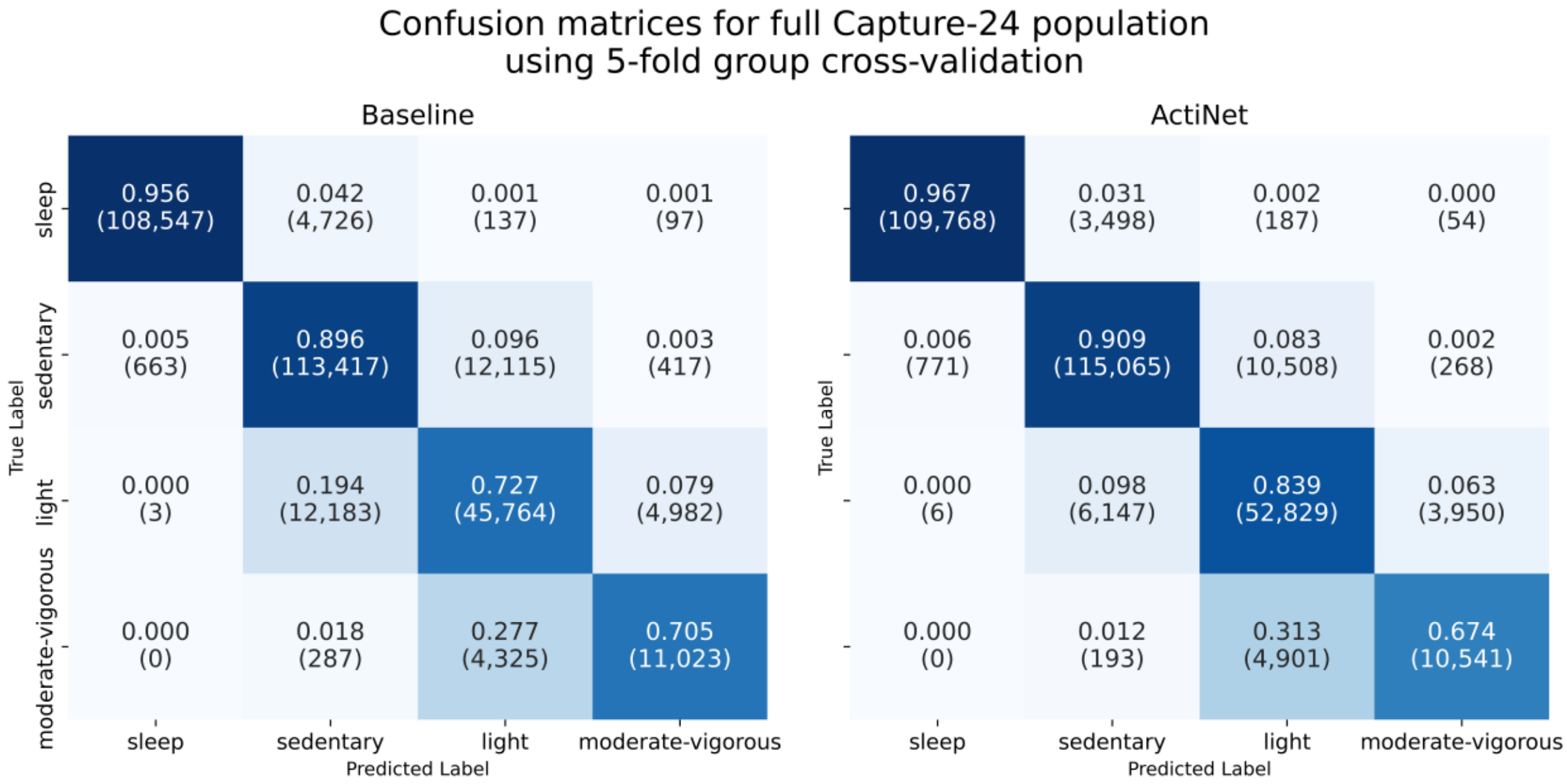

The confusion matrices presented in Figure 1 offer further insight into the performance of the model. In particular, the ActiNet model exceeds the baseline in the classification of light activity. The classification true positive rate for light activity in the ActiNet model was 0.84, as compared to the baseline at 0.73.

The model was then trained on the full CAPTURE-24 dataset, with data from 80% of participants used for training the HARNet model, and the remaining 20% reserved for early stopping and HMM training. The resulting HMM parameters are reported in the Supplementary HMM tables, which include the transition and emission matrices for the fully trained model. The final model was subsequently deployed to the open-source ActiNet repository, to be compared to the accelerometer pip package, a deployed version of the RF + HMM baseline model.

The baseline (Accelerometer) and ActiNet models were then run on the Realworld[42] and WISDM[43] datasets, to compare their performance. Table 2 shows the performance of the models on these datasets, with confusion matrices in Supplementary Figure 2. The ActiNet model outperforms the baseline model on the Realworld and WISDM datasets according to all evaluation metrics explored, boosting the mean macro F1 score from 0.53 to 0.66 and 0.68 to 0.74, respectively. The confusion matrices highlight the difficulties both models had in reliably detecting light activity in these non-free-living datasets.

*Table 2: Evaluation of the performance of the CAPTURE-24 trained activity recognition models. The baseline model is a random forest-based model, pip:accelerometer. ActiNet is a HARNet-based model, pip:actinet. Both models contain hidden Markov model smoothing, and post hoc sleep correction. Scores are reported as the mean ± standard deviation for participant scores for the evaluation on the Realworld[42] and WISDM[43] datasets. The highest score for a given dataset is indicated in bold.*

| Dataset | Model | Accuracy | Balanced Accuracy | Cohen's kappa | Macro F1 |
|---|---|---|---|---|---|
| Realworld | Baseline | 0.785 ± 0.074 | 0.605 ± 0.083 | 0.629 ± 0.114 | 0.531 ± 0.116 |
| | ActiNet | **0.813 ± 0.088** | **0.710 ± 0.149** | **0.684 ± 0.139** | **0.661 ± 0.150** |
| WISDM | Baseline | 0.755 ± 0.096 | 0.723 ± 0.091 | 0.621 ± 0.134 | 0.679 ± 0.118 |
| | ActiNet | **0.784 ± 0.080** | **0.759 ± 0.082** | **0.663 ± 0.124** | **0.741 ± 0.097** |

## Activity Composition

We also investigated how the proportion of classification outputs per participant differed between the two models and the ground-truth annotation. First, we investigated how the error in the predicted volume of each activity intensity averaged across each participant differed between the two models. These results are presented in Table 3 and reveal a significant reduction in activity volume prediction error, particularly in light activity and sedentary behaviour.

*Table 3: Mean absolute errors in hours for estimation of the total amount of each activity intensity in CAPTURE-24 population, comparing the baseline (random forest + hidden Markov model) and ActiNet model to ground-truth labels. Mean absolute error is measured in hours, with p-values reported for paired t-tests for errors in estimated activity time.*

| **Model** | **Mean absolute error in activity intensity classification [hours]** | | | |
|---|---|---|---|---|
| | **Sleep** | **Sedentary** | **Light** | **Moderate-vigorous** |
| **Baseline** | 0.299 ± 0.706 | 0.768 ± 0.823 | 0.783 ± 0.804 | 0.358 ± 0.560 |
| **ActiNet** | 0.240 ± 0.616 | 0.532 ± 0.713 | 0.581 ± 0.722 | 0.332 ± 0.522 |
| ***p-value*** | *0.004* | *<0.001* | *<0.001* | *0.013* |

Figure 2 further presents the distribution of these errors, using boxplots. The ActiNet model tended to predict the amount of time in each activity intensity closer to the true amount, indicated by boxplots closer to 0 hours. In addition, the precision of the prediction was also improved in ActiNet compared to the baseline random forest model, as the size of the ActiNet boxplots are smaller, indicating a lower spread in prediction error.

*Figure 2: Box plots for the distribution of errors in the estimation of the total amount of each activity intensity in CAPTURE-24 population, comparing the baseline and ActiNet model to ground-truth labels. The baseline model is a random forest-based model, and ActiNet is a HARNet-based model. Both models consist of hidden Markov model smoothing and post hoc sleep correction.*

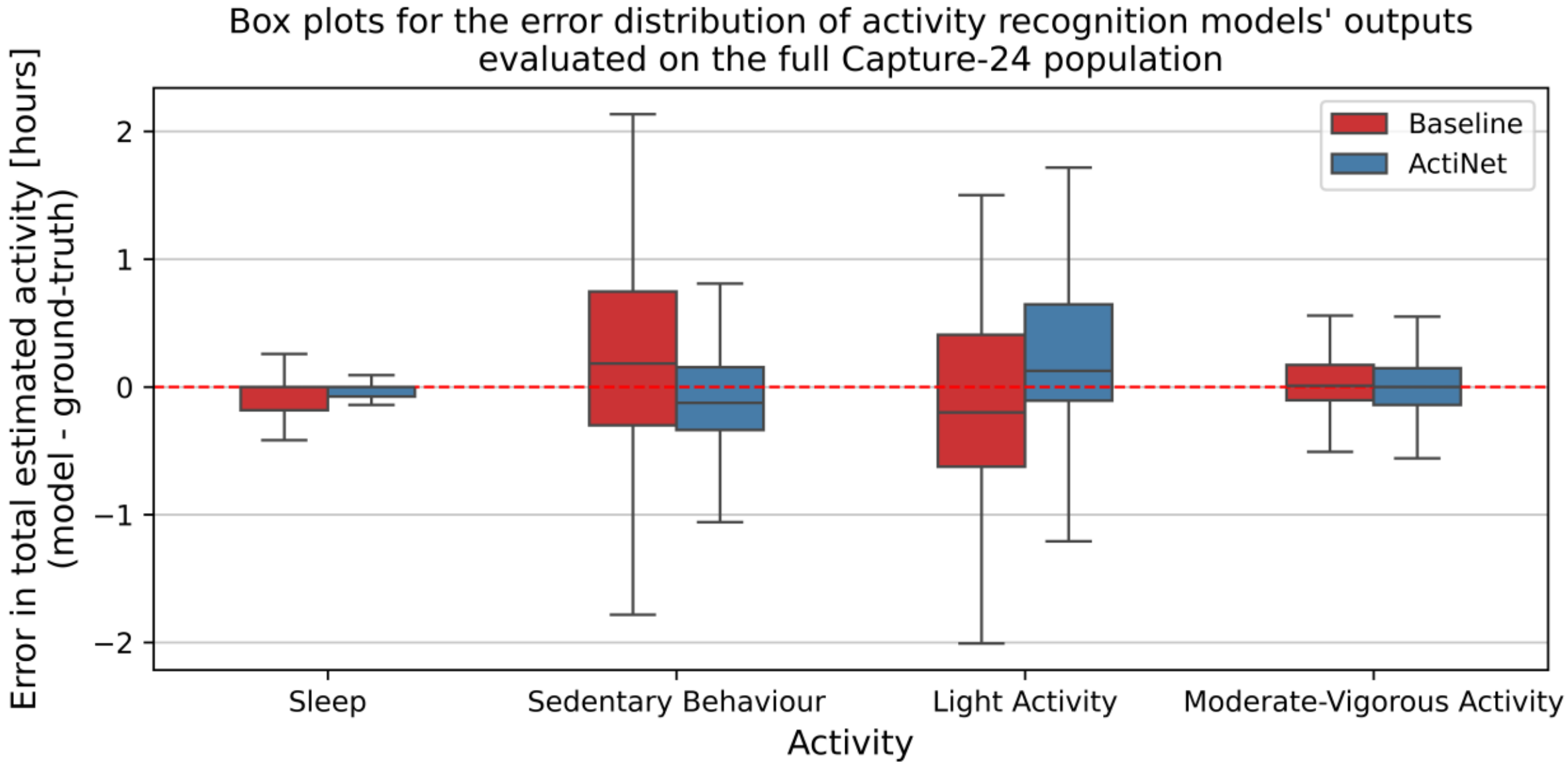

We then compared the total amount of each activity intensity, estimated by both models, and the ground-truth labels, using Bland-Altman plots and Pearson correlation, presented in Figure 3. These estimates typically showed very strong correlation (Pearson coefficient > 0.8)[45] between the different models and the ground-truth labels. Correlation coefficients were higher when comparing the ActiNet model to ground-truth labels, as compared to the baseline model to ground-truth labels, for each respective activity intensity.

In comparing the two models' prediction outputs, the mean differences tended to be around 0 for all output labels, but indicated a tendency for the ActiNet model, averaged across all CAPTURE-24 participants, to predict more sleep (0.07 hours) and light activity (0.34 hours) while classifying less sedentary behaviour (0.32 hours) and moderate-vigorous activity (0.09 hours), as compared to the baseline model. There was, however, a wide range in the upper and lower limits of agreement in activity intensity outputs, particularly for sedentary and light behaviour, indicating a greater degree of random, rather than systematic bias in the differences.

*Figure 3: Bland-Altman plots comparing the amount of activity intensity by a) Ground Truth and the baseline model b) Ground Truth and ActiNet model and c) Baseline and ActiNet models, evaluated on the CAPTURE-24 dataset using 5-fold group cross-validation. The baseline model is a random forest-based model, and ActiNet is a HARNet-based model. Both models consist of hidden Markov model smoothing and post hoc sleep correction. Each Bland–Altman plot shows the difference between two activity intensity estimates against their mean, with the mean bias (solid red line) and 95% limits of agreement (dashed blue lines) indicating agreement.*

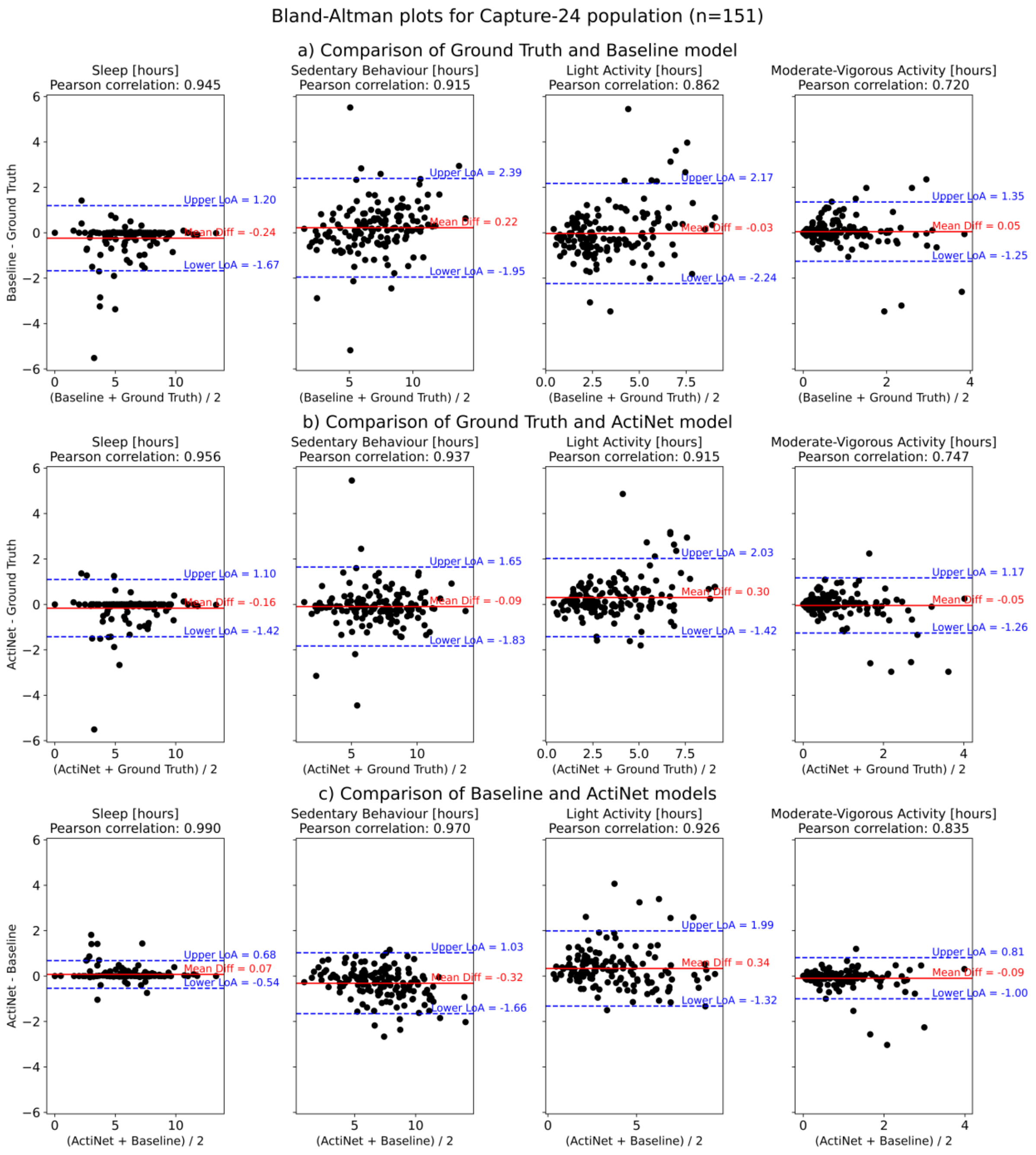

## Subgroup Patterns

We also investigated whether there were any noticeable differences in model performance, or classification outputs within subgroups of age or sex. Confusion matrices for the performance of the model within these subgroups are attached in the Supplementary materials, showing no noticeable difference in model performance within subgroups of age or sex. Furthermore, within these subgroups, the performance improvement of ActiNet over the baseline model, indicated using macro F1 score, is displayed using box plots in Figure 4. In this plot, we see that for all subgroups explored, the macro F1 scores for participants tended to be higher when using the ActiNet model.

*Figure 4: Performance of activity intensity classification models on the CAPTURE-24 dataset within subgroups of a) Age Band and b) Sex. Performance metrics reported using Macro F1 score. The baseline model is a random forest-based model, and ActiNet is a self-supervised modified ResNet-V2-based model. Both models consist of hidden Markov model smoothing and post hoc sleep correction.*

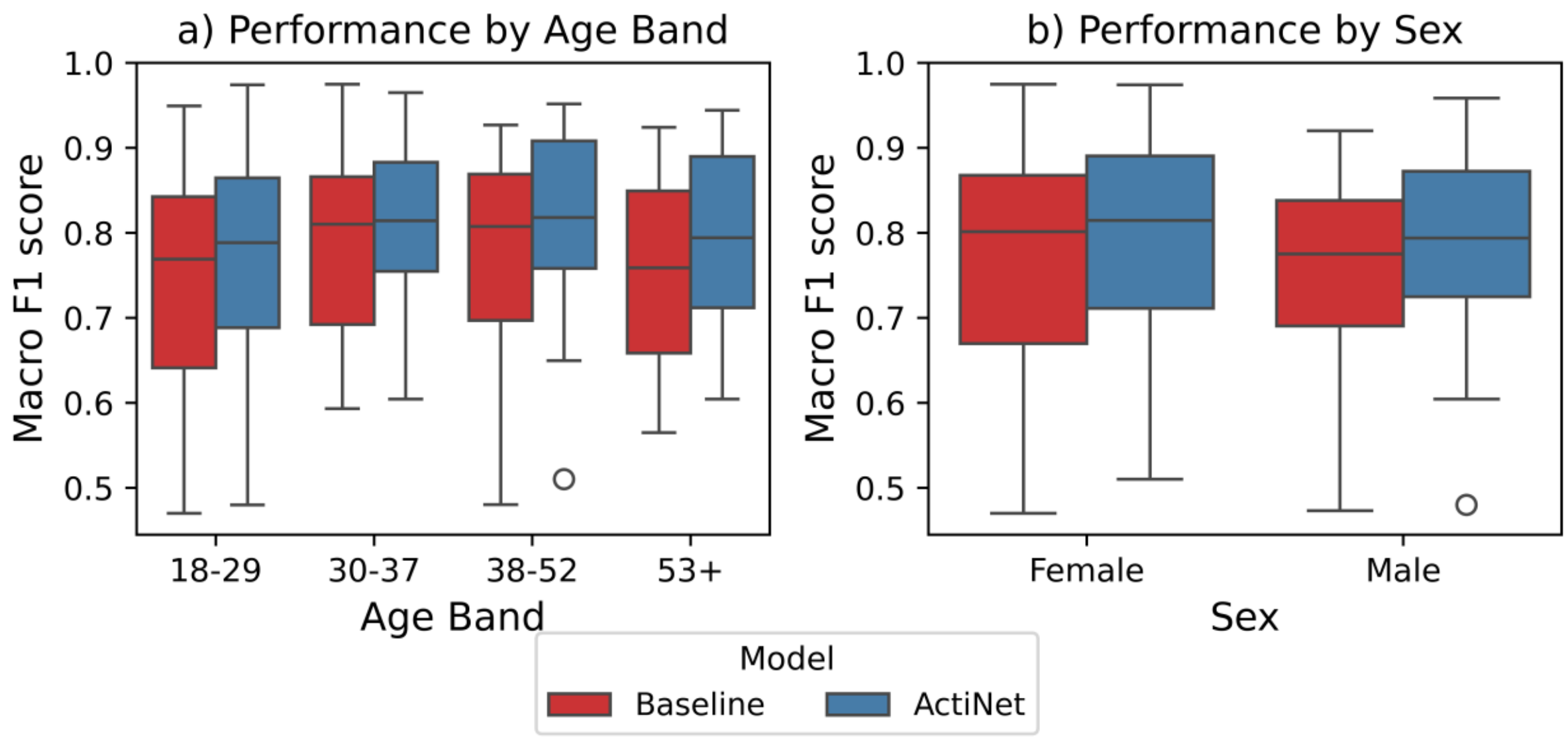

# Discussion

Based on the findings in this study, the ActiNet model improved the classification of labels of activity intensity (sleep, sedentary, light or moderate-to-vigorous activity) from wrist-worn accelerometry as compared to the baseline random forest + hidden Markov (HMM) model. The use of a self-supervised learning (SSL) modified ResNet-V2 model + HMM smoothing led to improved classification performance according to the per-participant averaged metrics evaluated in this study, consistent across subgroups of age and sex, as well as on the external validation datasets. The resulting outputs of the new ActiNet model maintained a very strong correlation with the baseline model in the aggregated output labels for each participant, however, on average, lower amounts of classified sedentary behaviour and moderate-to-vigorous physical activity were observed.

The improvement in classification performance through the adoption of the self-supervised model was consistent with that seen in prior literature[14,19,20]. The deep learned features extracted using the HARNet model provided greater variance for separating different activity intensity labels, over the handcrafted features used by the accelerometer model, making it easier for the model to differentiate activity intensity[14]. The improvements in classification performance were maintained even after the addition of HMM smoothing and sleep block correction, removing transient classification outputs by adding temporal dependencies. This is because, while these post-processing steps can alter the predictions made, the predictions are largely driven by the core HARNet or RF model.

Classification models had higher true positive rates in the classification of sleep and sedentary behaviour, as compared to light and moderate-vigorous activity. Light activity could either be misclassified as sedentary, due to immobile standing periods, or as moderate-vigorous activity, due to active periods that do not quite cross the 3 METs threshold[23]. Both models struggled to reliably detect light activity in the external validation datasets. These datasets were collected in non-free-living settings, resulting in less natural patterns in light activity behaviour, potentially limiting the effectiveness of the HMMs.

These findings may have impacts on large-scale epidemiology studies. The biobank accelerometer analysis tool[10], for example, has been used in existing studies, including Walmsley et al. (2022)[1], Shreves et al. (2025)[2] and Le et al. (2026)[3], to observe health associations with activity levels. The results for this study suggest that we may be able to better capture the volume of each activity intensity performed by participants, which may, in turn, strengthen the observed associations by reducing random error and therefore regression dilution bias. This may help increase statistical power for health association studies and provide better estimates for public health guidelines. Improved classification accuracy may also enable the reliable extraction of additional digital measures of activity, such as bout-related characteristics, rather than being limited to total daily activity volume.

Important considerations should however be made with the adoption of deep learning models in future machine learning (ML)-based activity intensity recognition, in accordance with ethical research principal frameworks, like the European Union Artificial Intelligence (EU AI) Act[46,47]. Despite the performance improvements noted in this study, deep learning models come coupled with some additional considerations. Firstly, the deep learned features, extracted by the model have less explainability for model operation and failure, as compared to the hand-crafted features used for random forest models, or simpler thresholds of average acceleration in cutpoint-based models[48]. Furthermore, deep learning models such as ActiNet are more complex than traditional ML approaches like the baseline model, hence require greater storage size, and typically demand more time for training and deployment. These deep-learning models also require a graphical processing unit (GPU) for training, resulting in a larger carbon footprint[49].

However, the use of a pre-trained model reduced computational demand and associated carbon footprint by reusing existing weights, accelerating training compared to training from scratch. Our use of early stopping after five training epochs without validation improvement further limited unnecessary computation. Making the model open-source and publicly available also reduces redundant retraining by enabling shared, standardised outputs across research teams, improving overall efficiency and sustainability.

The study has many strengths supporting its findings. We used the CAPTURE-24 dataset, which is a very large, comprehensive dataset, capturing a wide array of data in free-living. This dataset, as well as the code used to produce these results, have been made publicly available for purposes of reproducibility and iterative improvement. We have made use of modern ML models and techniques while building on existing baseline models used in prior research. Finally, the use of external validation datasets, as well as an investigation into subgroups of age and sex, gives further validity to the findings.

There were also limitations to the study. Firstly, the external validation datasets were much smaller in size compared to the training dataset and consisted only of daytime recordings collected in supervised laboratory settings. Consequently, these datasets excluded sleep data and may not fully reflect the variability and complexity of activities performed in free-living environments, limiting our ability to assess the model's generalisability across other datasets, and the wider population. The use of HMMs, while effective in this study, were restricted to training on only 20% of the training data, and hence may underperform as compared to the RF-based HMM. The CAPTURE-24 training data also has some limitations, due to a lack of diversity in tasks performed by the participants, all from the Oxfordshire area, limiting its generalisability to all populations across the globe. This dataset also relies on the use of sleep diaries and camera images taken every 20-40 seconds for ground-truth annotations, hence may miss some critical activity behaviour from invalid diaries entries, or between camera images[32]. Our use of deep learning also limited the interpretability of model operation, as compared to traditional ML approaches, which is of great importance in health-related research.

Future work should take steps to address some of the considerations with deploying deep learning models, including strategies to reduce carbon footprint, model size, and computational cost during training and inference. Explainable artificial intelligence (XAI) techniques and ablation studies could also be applied to interpret the learned representations and to provide greater insight into model behaviour. Further investigations into failure cases of the model, such as low sampling frequency, or data with high rates of clipping, would offer insight into the reliability of the model on other datasets. In addition, deep sequence learning architectures, such as long short-term memory (LSTM) networks or transformer-based models, could be explored, although such approaches may require larger annotated datasets than CAPTURE-24. The ActiNet model could also be extended to classify activities of daily living, including behaviours such as walking and cycling, thereby enabling a more detailed characterisation of human activity, or alternatively be trained and deployed on other wear locations, such as the thigh, rather than just the wrist. Finally, future studies could investigate how the adoption of ActiNet, relative to existing models like the accelerometer pip package, influences observed associations with health outcomes in large-scale cohorts such as UK Biobank.

# Conclusion

This study introduced ActiNet, a novel approach to activity intensity recognition, using self-supervised deep learning, with hidden Markov model smoothing. The results from this study demonstrate that ActiNet outperforms a baseline random forest-based model. These findings were consistent across subgroups of age and sex. This encourages the use of ActiNet in future epidemiological studies, extracting labels of activity intensity from wrist-worn accelerometer data.

# Code and Data Availability

The CAPTURE-24 dataset used in this analysis is available at https://ora.ox.ac.uk/objects/uuid:99d7c092-d865-4a19-b096-cc16440cd001.

The code used to process this data, train the ActiNet and Random Forest models, and evaluate these models is available at https://github.com/OxWearables/actinet, and can be used free of charge for non-commercial purposes.

# Funding

This work was supported by SwissRe (AA, AD); Novo Nordisk (SC, AD); the British Heart Foundation Centre of Research Excellence [RE/18/3/34214] (AD); and Wellcome Trust [223100/Z/21/Z] (AD).

# Authors and Affiliations

1. **Big Data Institute, University of Oxford, Oxford, UK**

Aidan Acquah, Shing Chan & Aiden Doherty

2. **Nuffield Department of Population Health, University of Oxford, Oxford, UK**

Aidan Acquah, Shing Chan & Aiden Doherty

## Contributions

All authors (A.A., S.C., and A.D.) conceived the experiments, A.A. and S.C. conducted the experiments and analysed the results. A.A. wrote the first draft of the manuscript. All authors (A.A., S.C., and A.D.) reviewed the manuscript.

## Corresponding Author

Correspondence to Aidan Acquah (a.acquah@ucl.ac.uk).

# Acknowledgements

We thank Alyona Tsvetkova for providing feedback on the evaluation of the model.

# Ethics Declarations

## Competing Interests

The authors declare no competing interests.

## Ethics Approval and Consent to Participate

The ethical approval for the CAPTURE-24 dataset was granted by the University of Oxford Inter-Divisional Research Ethics Committee (Ref SSD/CUREC1A/13–262).

# Supplementary Materials

## Tables

*Supplementary Table 1: List of hand-crafted features extracted by the Walmsley accelerometer[1] pip package for the baseline random forest[2] model. ENMO is the Euclidean Norm Minus One[3]. g is acceleration due to gravity. v is the Euclidean norm of the signal in x, y and z. nats are units of information (e.g., entropy), Hz is frequency per second, and s is seconds.*

| **Feature Name** | **Description** | **Unit** |
|---|---|---|
| mean | Mean value of v | g |
| sd | Standard deviation of v | g |
| min, max | Minimum and maximum of v | g |
| 25thp, median, 75thp | 25th, 50th, 75th percentile of v | g |
| coefvariation | Coefficient of variation of v | |
| MAD | Mean absolute deviation of v | g |
| MPD | Mean pairwise difference of v | g |
| skew, kurt | Skewness, Kurtosis of v | |
| enmoTrunc, enmoAbs | Truncated, Absolute ENMO | g |
| autocorr | Auto-correlation of v | |
| fmax | Maximum frequency | Hz |
| pmax | Maximum power | $g^2/s$ |
| fmaxband | Frequency band of maximum power | Hz |
| pmaxband | Power of maximum frequency band | $g^2/s$ |
| entropy | Spectral entropy | nats |
| fft1, fft2, ..., fft10 | Power spectral density for frequencies 1Hz, 2Hz, ..., 10Hz | $g^2/s$ |
| f1, f2 | 1st and 2nd dominant frequencies | Hz |
| p1, p2 | Power spectral densities of respective dominant frequencies | $g^2/s$ |
| f625 | Dominant frequency between 0.6 and 2.5Hz | Hz |
| p625 | Power at dominant frequency between 0.6 and 2.5 Hz | $g^2/s$ |
| totalPower | Total power | $g^2/s$ |

*Supplementary Table 2: Test participant IDs for each fold in the cross-validation training and evaluation on the CAPTURE-24[4] dataset in this study. The remaining participant IDs (P001-P151) not listed as test IDs for the respective fold were used for model training, preventing data leakage.*

| Fold | Test Participant IDs |
|---|---|
| 1 | P001, P003, P006, P010, P012, P013, P017, P025, P030, P044, P056, P057, P059, P065, P070, P075, P079, P080, P082, P090, P092, P114, P115, P116, P117, P130, P131, P142, P145 |
| 2 | P004, P008, P014, P015, P016, P021, P023, P024, P028, P029, P032, P035, P037, P038, P043, P047, P048, P050, P074, P081, P085, P093, P094, P098, P099, P112, P118, P121, P127, P128, P146 |
| 3 | P026, P036, P039, P040, P042, P045, P046, P049, P052, P055, P064, P068, P069, P071, P072, P088, P091, P095, P100, P101, P103, P106, P109, P111, P126, P141, P144, P148, P149, P150, P151 |
| 4 | P002, P009, P011, P018, P019, P027, P031, P051, P053, P054, P060, P061, P067, P076, P077, P078, P083, P086, P108, P110, P113, P119, P122, P123, P132, P133, P135, P136, P138, P139, P147 |
| 5 | P005, P007, P020, P022, P033, P034, P041, P058, P062, P063, P066, P073, P084, P087, P089, P096, P097, P102, P104, P105, P107, P120, P124, P125, P129, P134, P137, P140, P143 |

# Hidden Markov models

Below shows the transition, and emission matrices from the training of ActiNet model on the full CAPTURE-24 population. Transition matrices represent the probability of moving from a state **S** at time **t** to a state at time **t+1**. Emission matrices represent the probability of observing an activity intensity **O** at time **t** given the state **S** at time **t**. A state **S** and corresponding observation **O** respectively represent the actual and predicted activity intensity label: sleep, sedentary, light or moderate-vigorous activity (MVPA), for a 30-second time window. These models assume a uniform prior distribution.

Transition matrix:

| S(t) \ S(t+1) | Sleep | Sedentary | Light | MVPA |
|---|---|---|---|---|
| Sleep | 0.999 | 0.001 | 0.000 | 0.000 |
| Sedentary | 0.001 | 0.986 | 0.012 | 0.000 |
| Light | 0.000 | 0.028 | 0.963 | 0.009 |
| MVPA | 0.000 | 0.003 | 0.025 | 0.972 |

Emission matrix:

| S(t) \ O(t) | Sleep | Sedentary | Light | MVPA |
|---|---|---|---|---|
| Sleep | 0.861 | 0.126 | 0.010 | 0.003 |
| Sedentary | 0.085 | 0.700 | 0.153 | 0.062 |
| Light | 0.017 | 0.202 | 0.514 | 0.266 |
| MVPA | 0.003 | 0.036 | 0.166 | 0.795 |

# Figures

*Supplementary Figure 1: Illustration of nested cross-validation, for the training and evaluation of the activity recognition models on the CAPTURE-24 dataset. K outer loop splits are a 5-fold stratified group cross validation. The one inner split is a group shuffle split with random state of 42.*

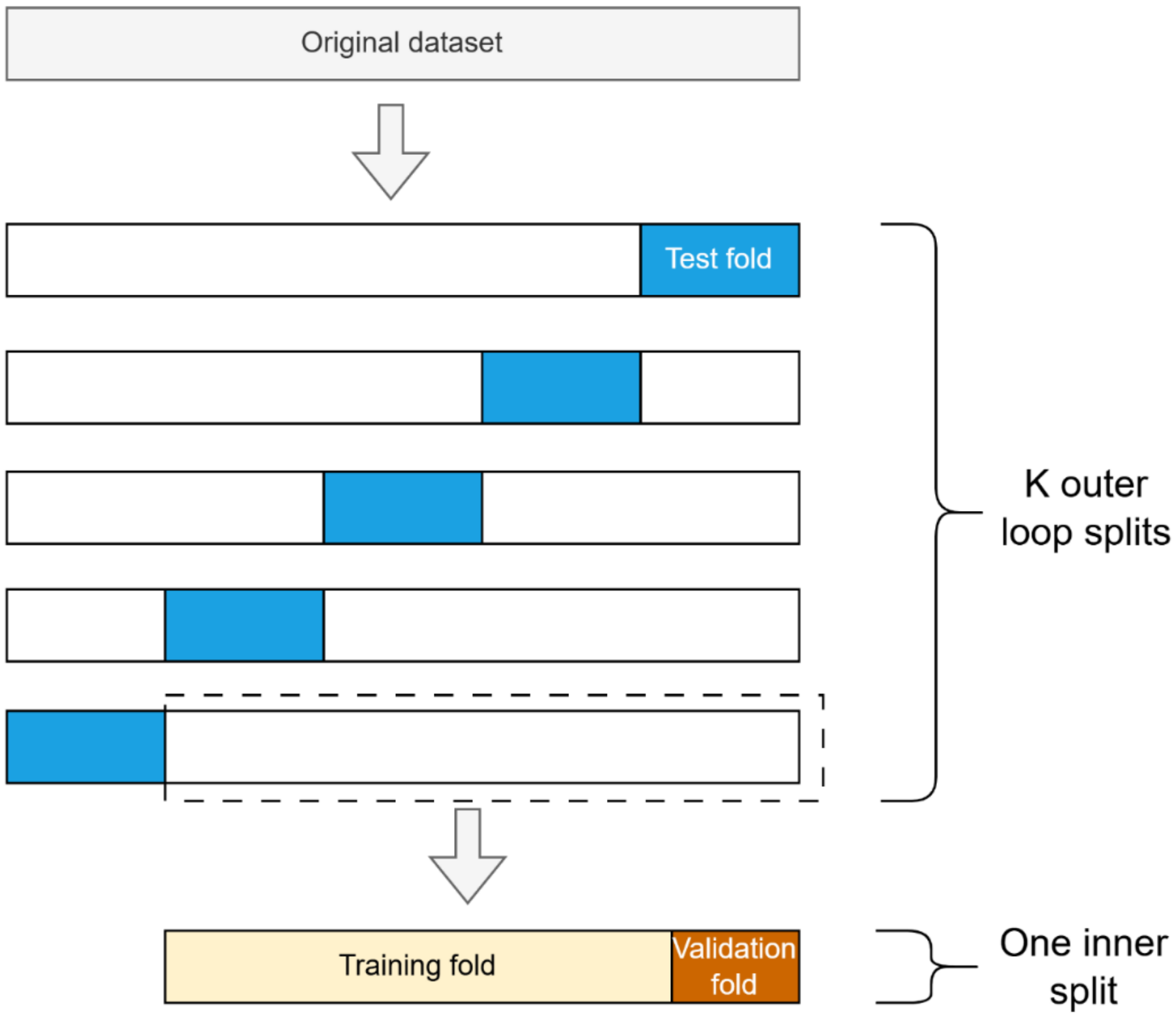

*Supplementary Figure 2: Confusion matrices for the performance of CAPTURE-24 trained activity recognition models, tested on the Realworld dataset (n = 14). The baseline model is a random forest-based model, ActiNet is a self-supervised HARNet-based model. Each cell shows the proportion of 30-second windows with a given true activity intensity label that were predicted as each activity intensity label, with corresponding window counts in parentheses.*

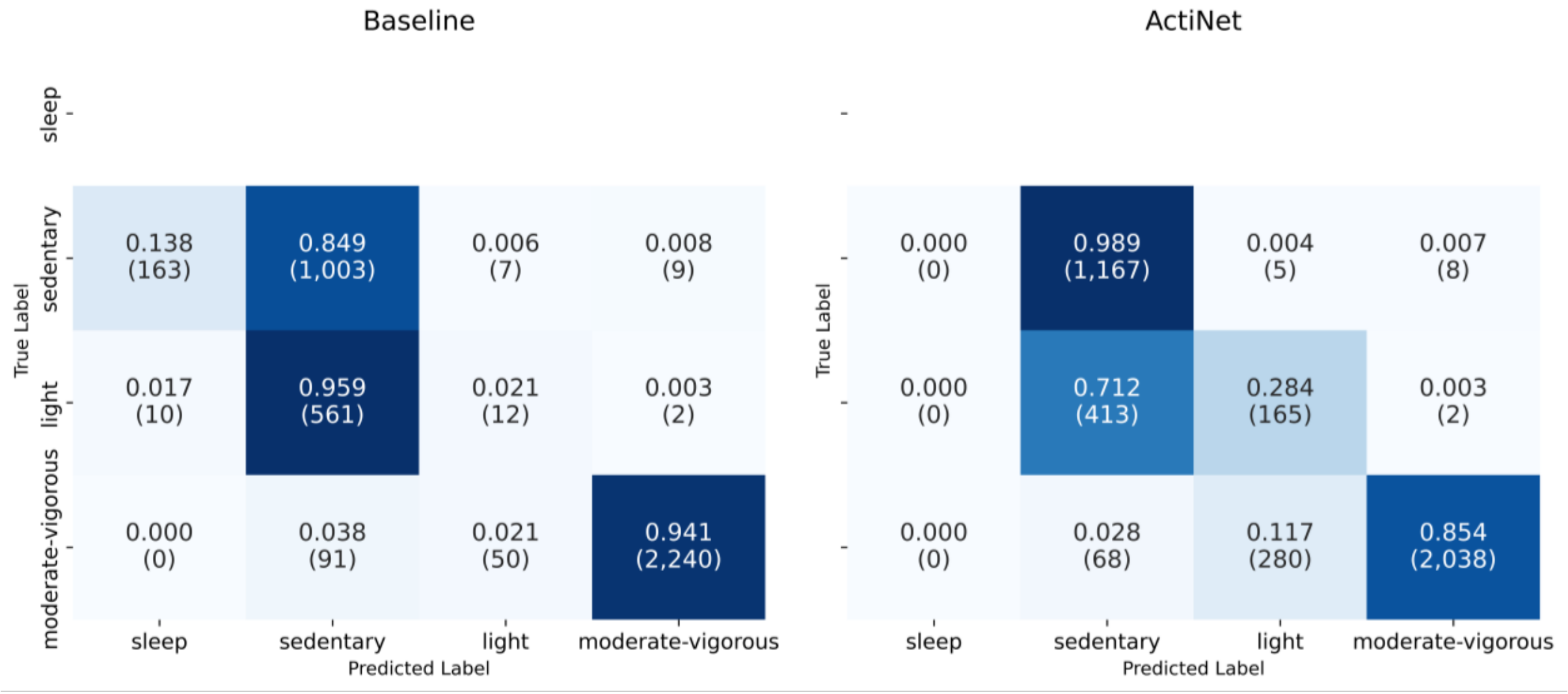

*Supplementary Figure 3: Confusion matrices for the performance of CAPTURE-24 trained activity recognition models, tested on the WISDM dataset (n = 46). The baseline model is a random forest-based model, ActiNet is a self-supervised HARNet-based model. Each cell shows the proportion of 30-second windows with a given true activity intensity label that were predicted as each activity intensity label, with corresponding window counts in parentheses.*

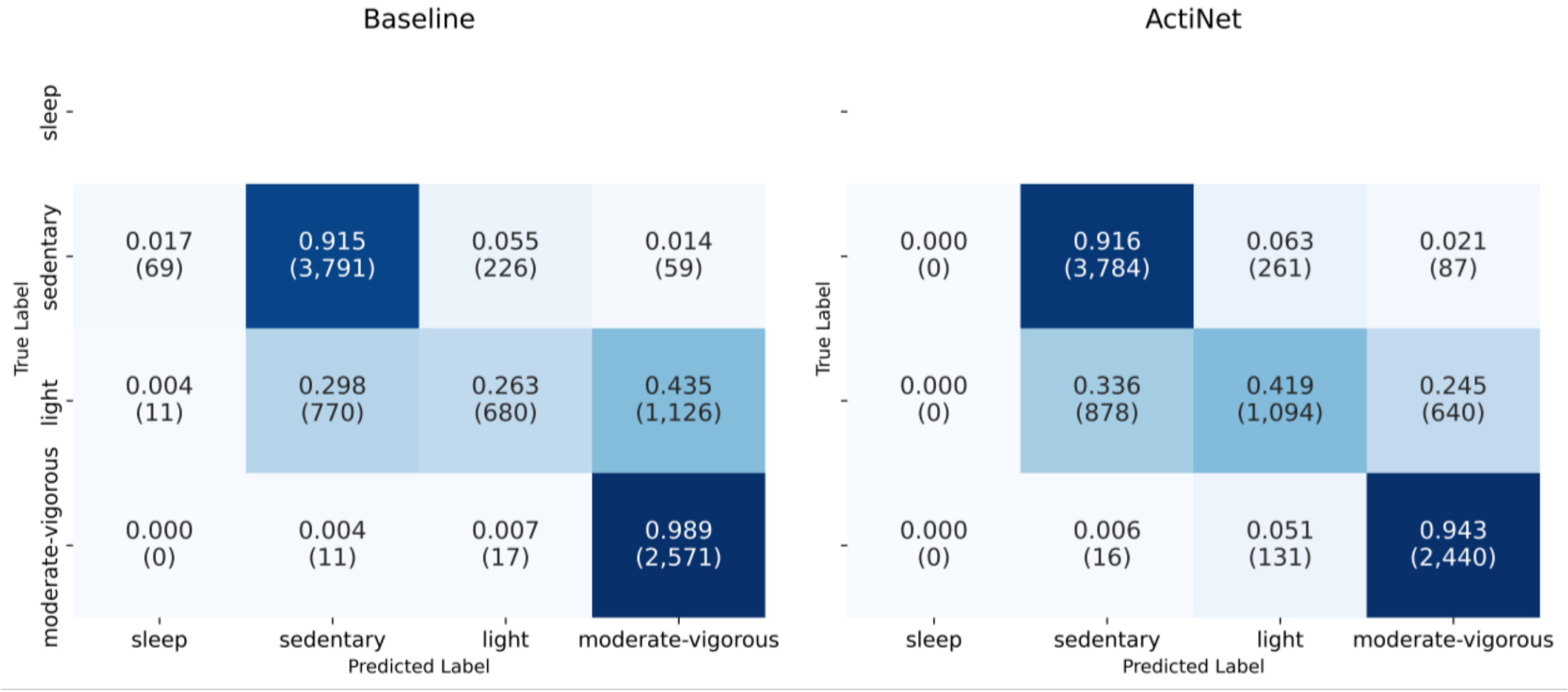

*Supplementary Figure 4: Confusion matrices for activity intensity classifications made by baseline and ActiNet models using 5-fold cross validation on CAPTURE-24 dataset, evaluated in age bands: 18-29, 30-37, 38-52 and 53+. Each cell shows the proportion of 30-second windows with a given true activity intensity label that were predicted as each activity intensity label, with corresponding window counts in parentheses.*

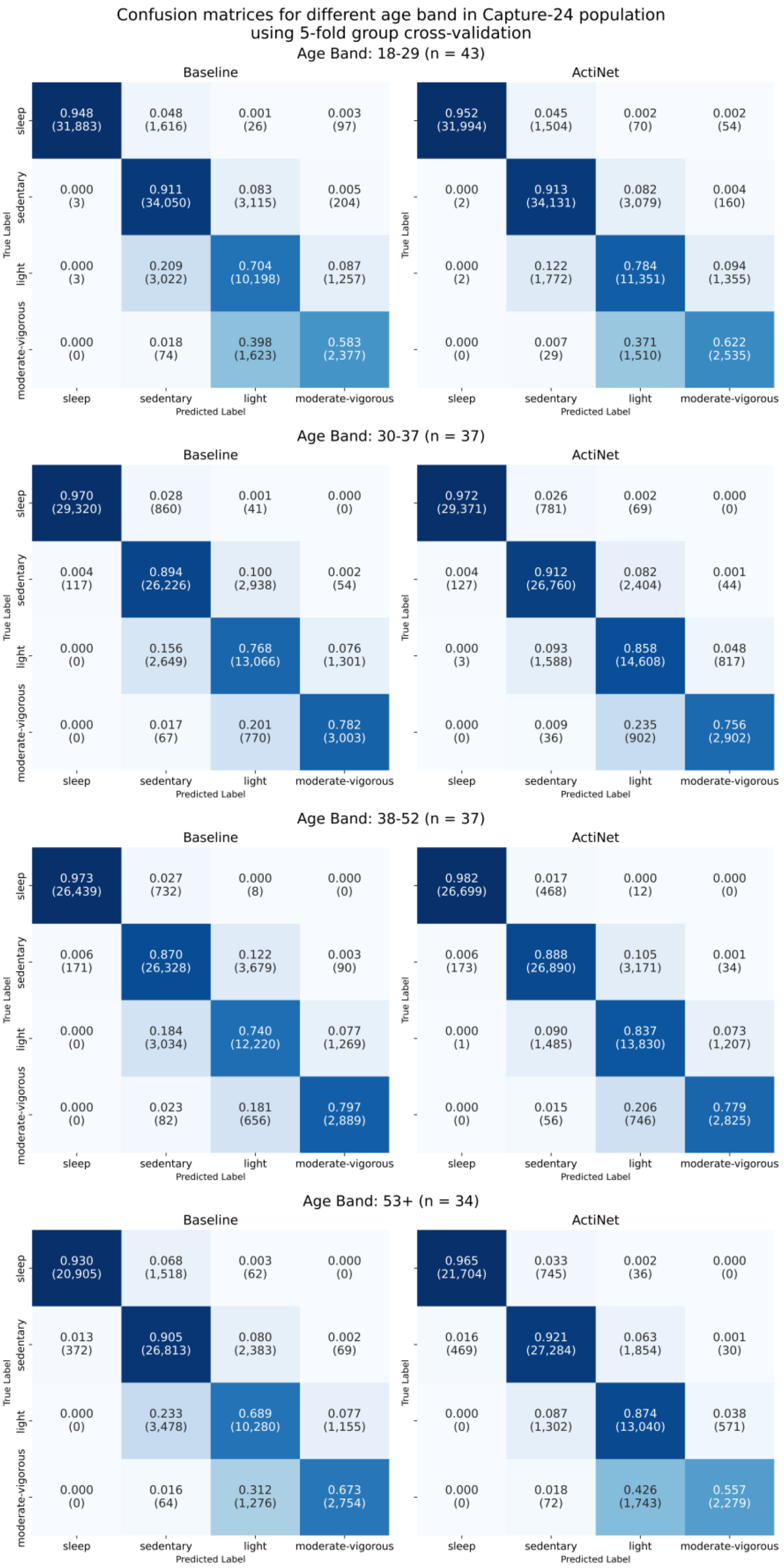

*Supplementary Figure 5: Confusion matrices for activity intensity classifications made by baseline and ActiNet models using 5-fold cross validation on CAPTURE-24 dataset, evaluated for different sexes: female and male. Each cell shows the proportion of 30-second windows with a given true activity intensity label that were predicted as each activity intensity label, with corresponding window counts in parentheses.*

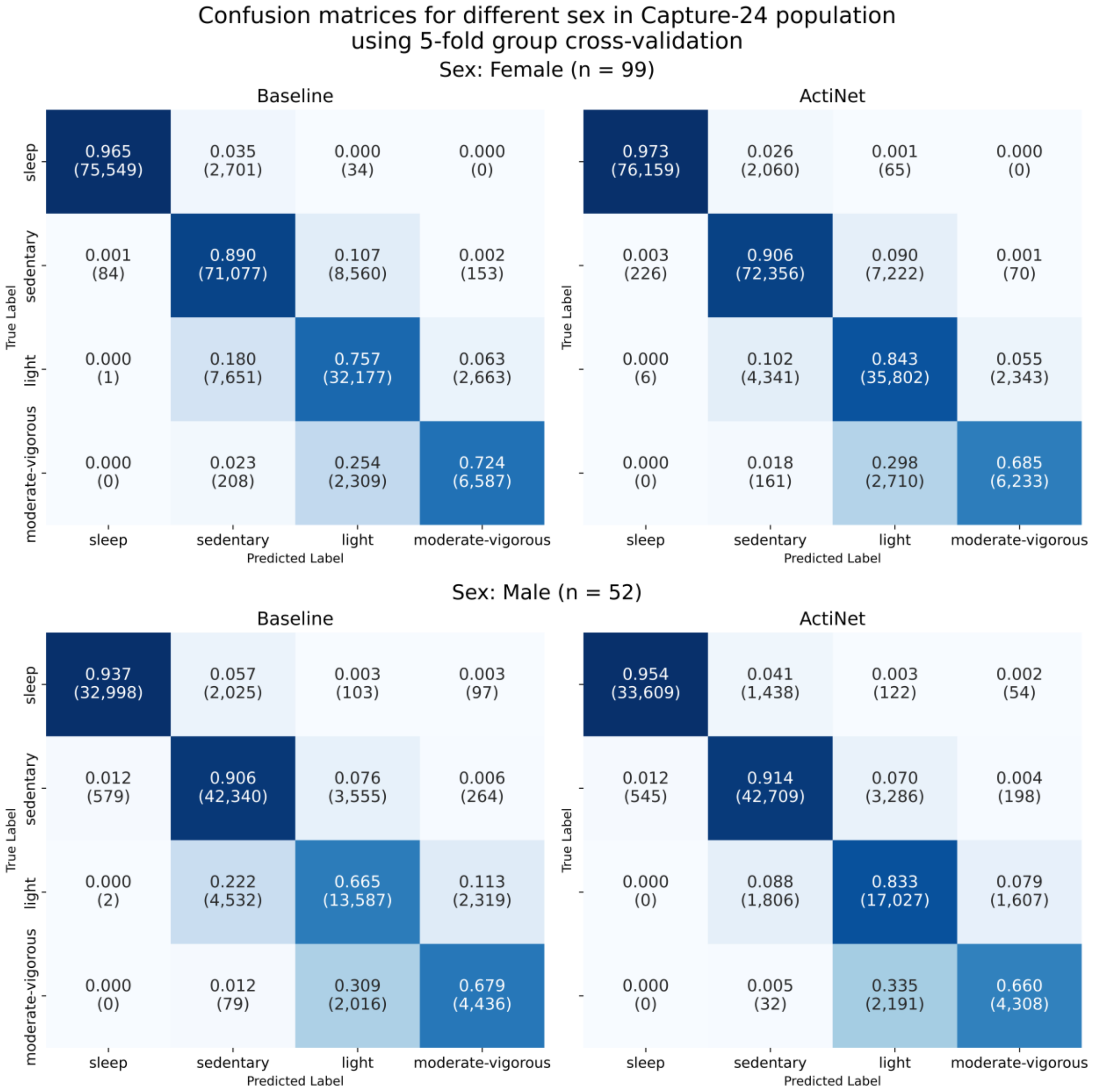

*Supplementary Figure 6: Box plots for the distribution of errors in total activity intensity estimation by the baseline and ActiNet models, evaluated on different age bands of the CAPTURE-24 dataset using 5-fold group cross-validation.*

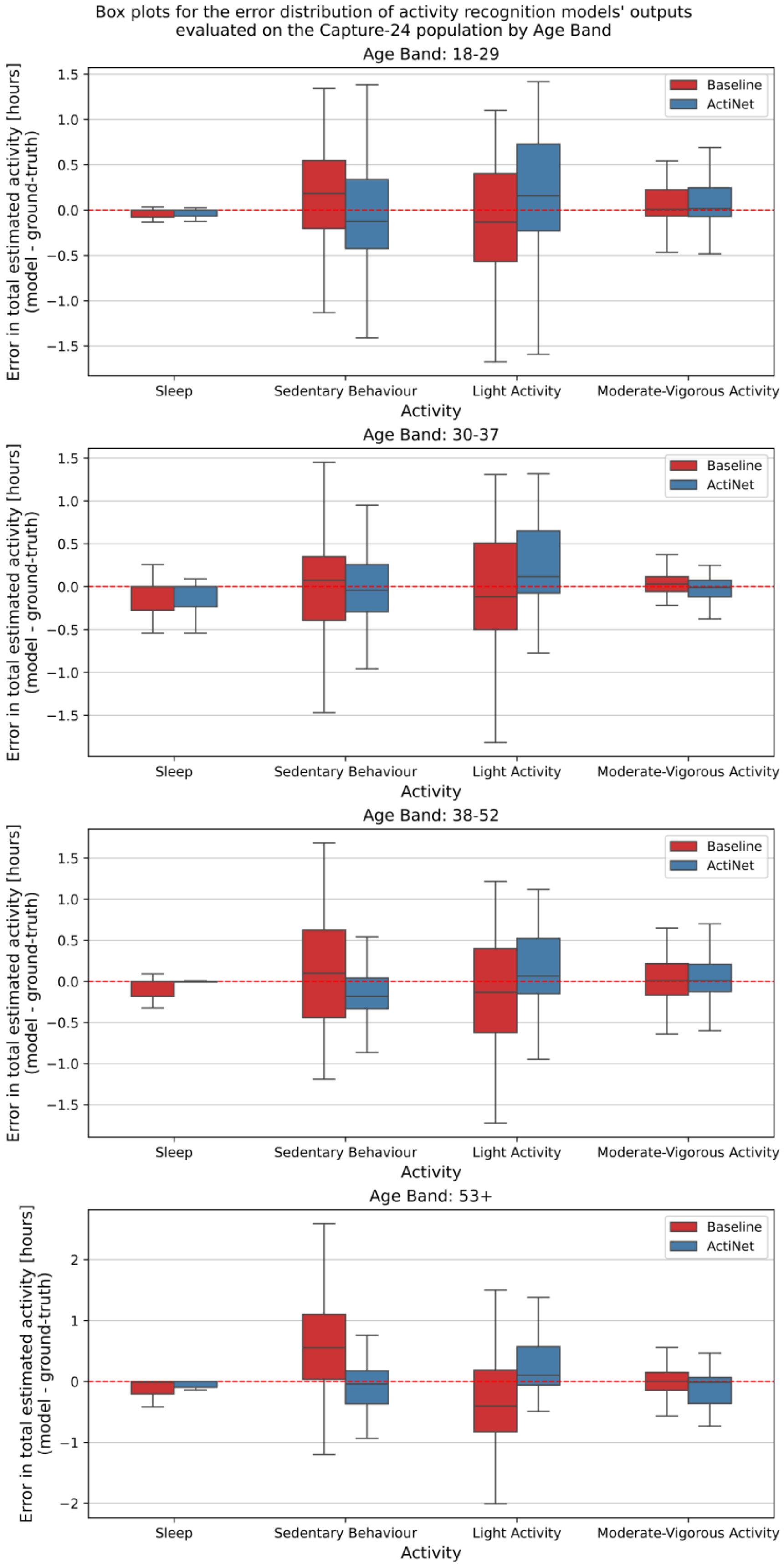

*Supplementary Figure 7: Box plots for the error in total activity intensity estimation by the baseline and ActiNet models, evaluated on different sexes of the CAPTURE-24 dataset using 5-fold group cross-validation.*

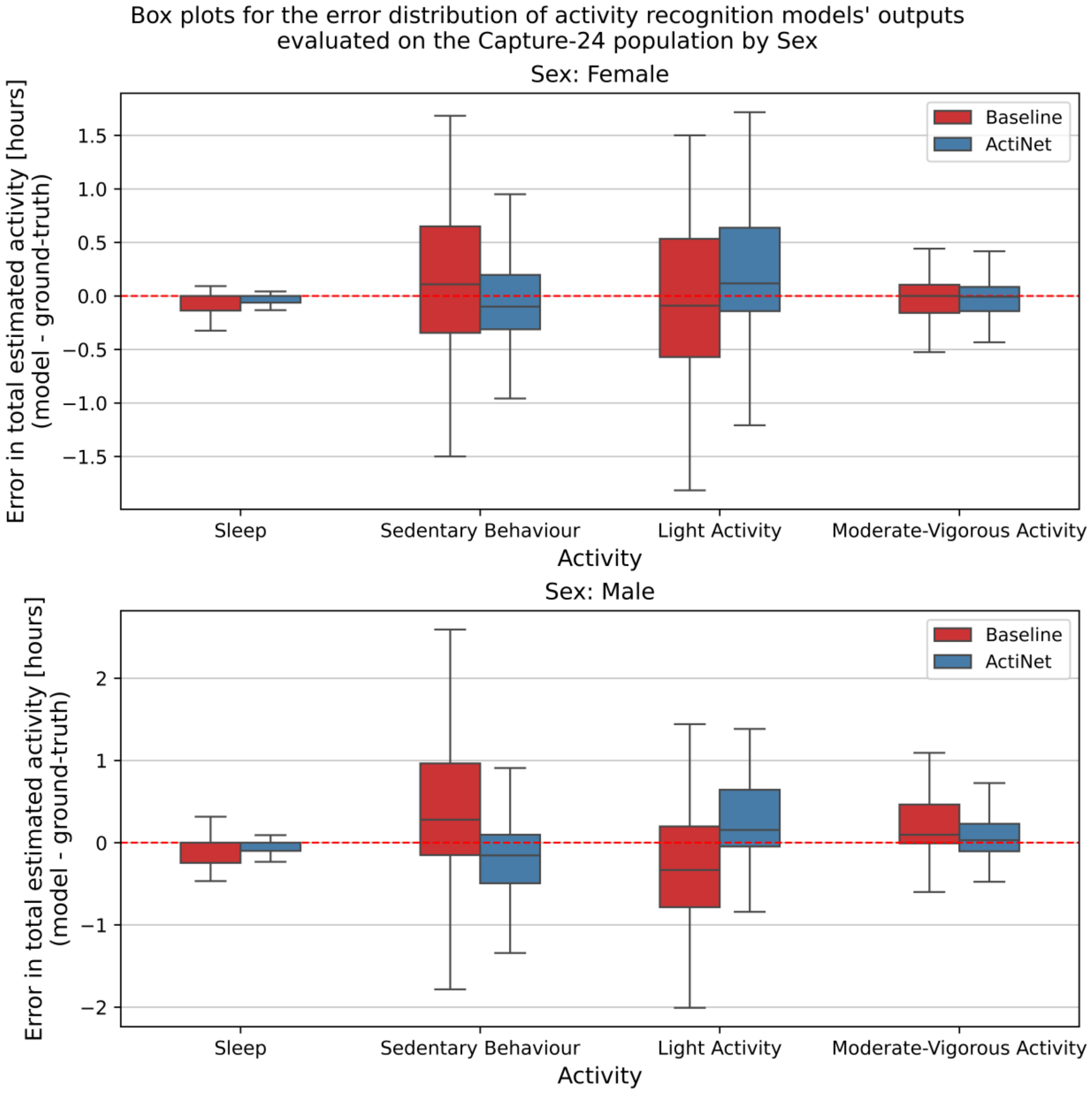

*Supplementary Figure 8: Bland-Altman plots comparing the amount of activity intensity from the ground-truth annotations to the classifications by the baseline random forest-based model, evaluated on different age bands of the CAPTURE-24 dataset using 5-fold group cross-validation. Each Bland–Altman plot shows the difference between two activity intensity estimates against their mean, with the mean bias (solid red line) and 95% limits of agreement (dashed blue lines) indicating agreement.*

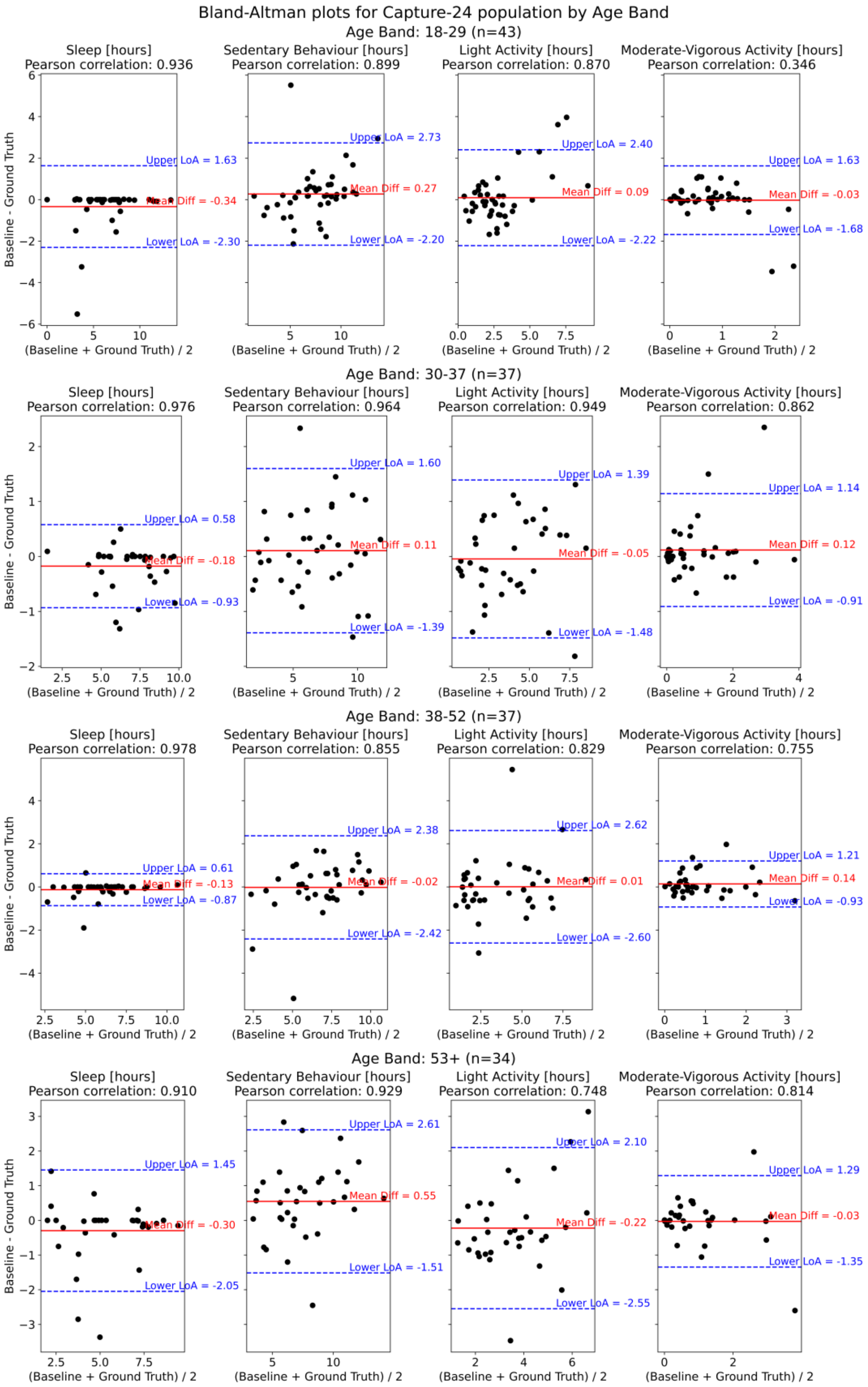

*Supplementary Figure 9: Bland-Altman plots comparing the amount of activity intensity from the ground-truth annotations to the classifications by the baseline random forest-based model, evaluated on different sexes of the CAPTURE-24 dataset using 5-fold group cross-validation. Each Bland–Altman plot shows the difference between two activity intensity estimates against their mean, with the mean bias (solid red line) and 95% limits of agreement (dashed blue lines) indicating agreement.*

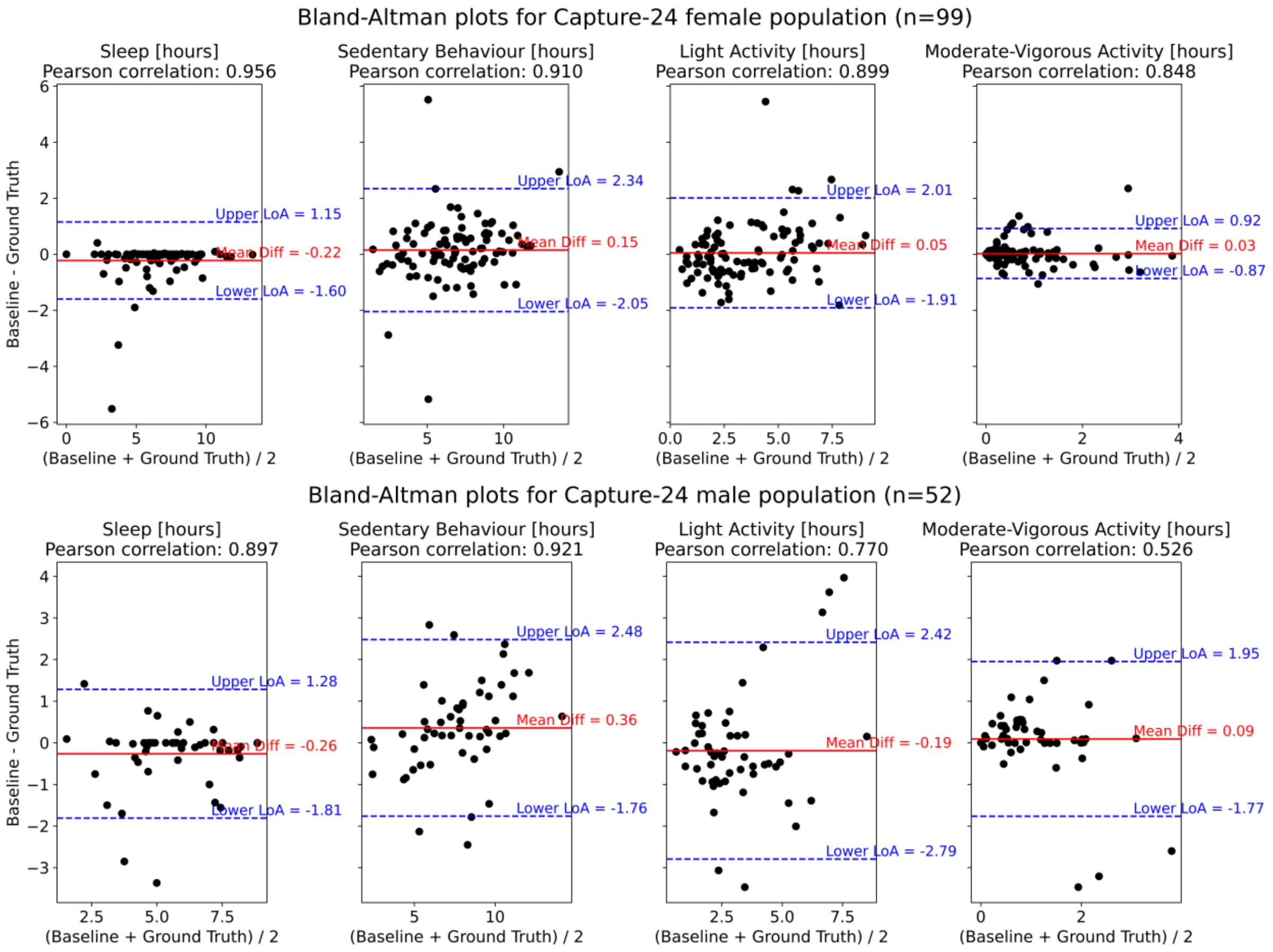

*Supplementary Figure 10: Bland-Altman plots comparing the amount of activity intensity from the ground-truth annotations to the classifications by the ActiNet model, evaluated on different age bands of the CAPTURE-24 dataset using 5-fold group cross-validation. Each Bland–Altman plot shows the difference between two activity intensity estimates against their mean, with the mean bias (solid red line) and 95% limits of agreement (dashed blue lines) indicating agreement.*

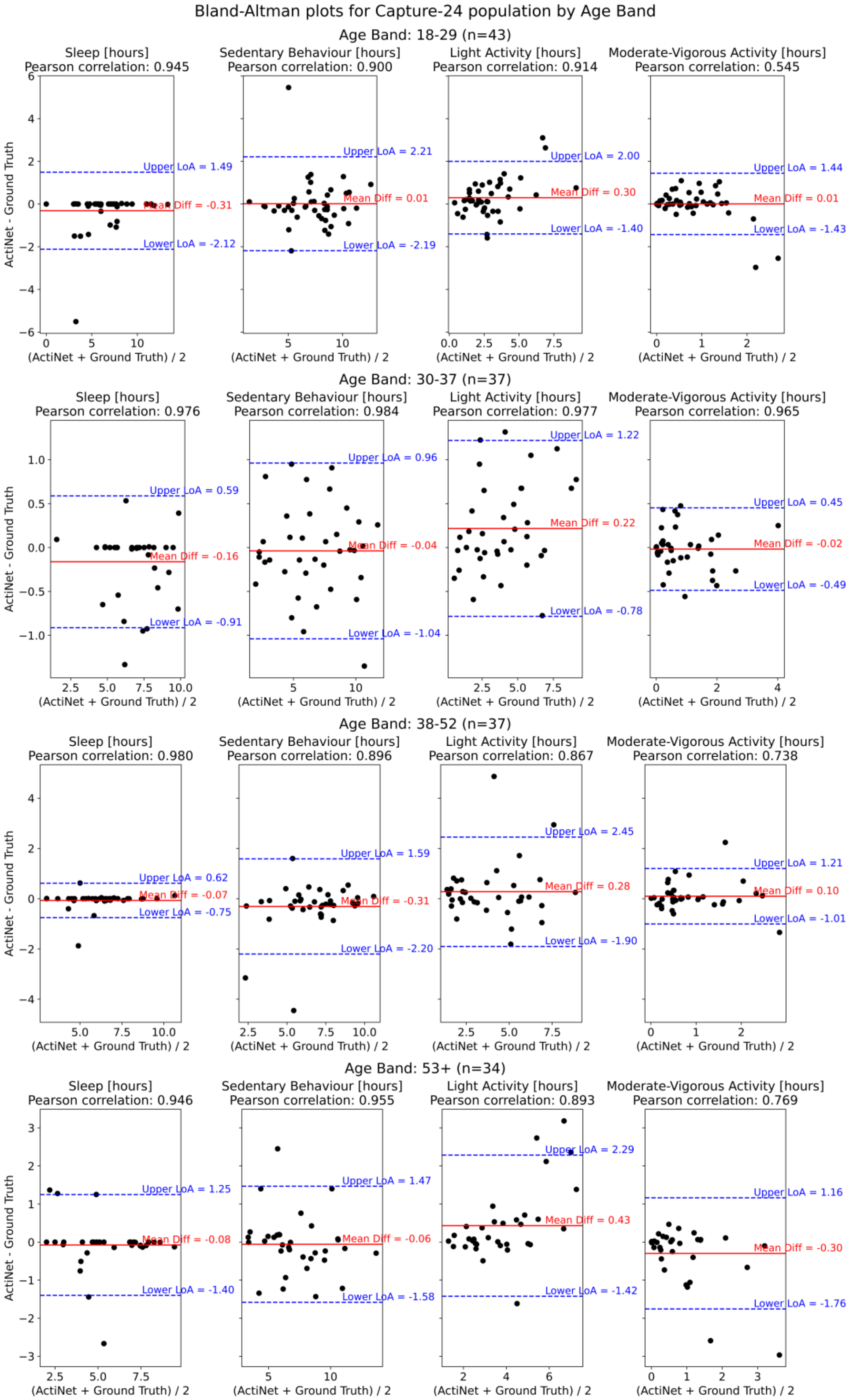

*Supplementary Figure 11: Bland-Altman plots comparing the amount of activity intensity from the ground-truth annotations to the classifications by the ActiNet model, evaluated on different sexes of the CAPTURE-24 dataset using 5-fold group cross-validation. Each Bland–Altman plot shows the difference between two activity intensity estimates against their mean, with the mean bias (solid red line) and 95% limits of agreement (dashed blue lines) indicating agreement.*

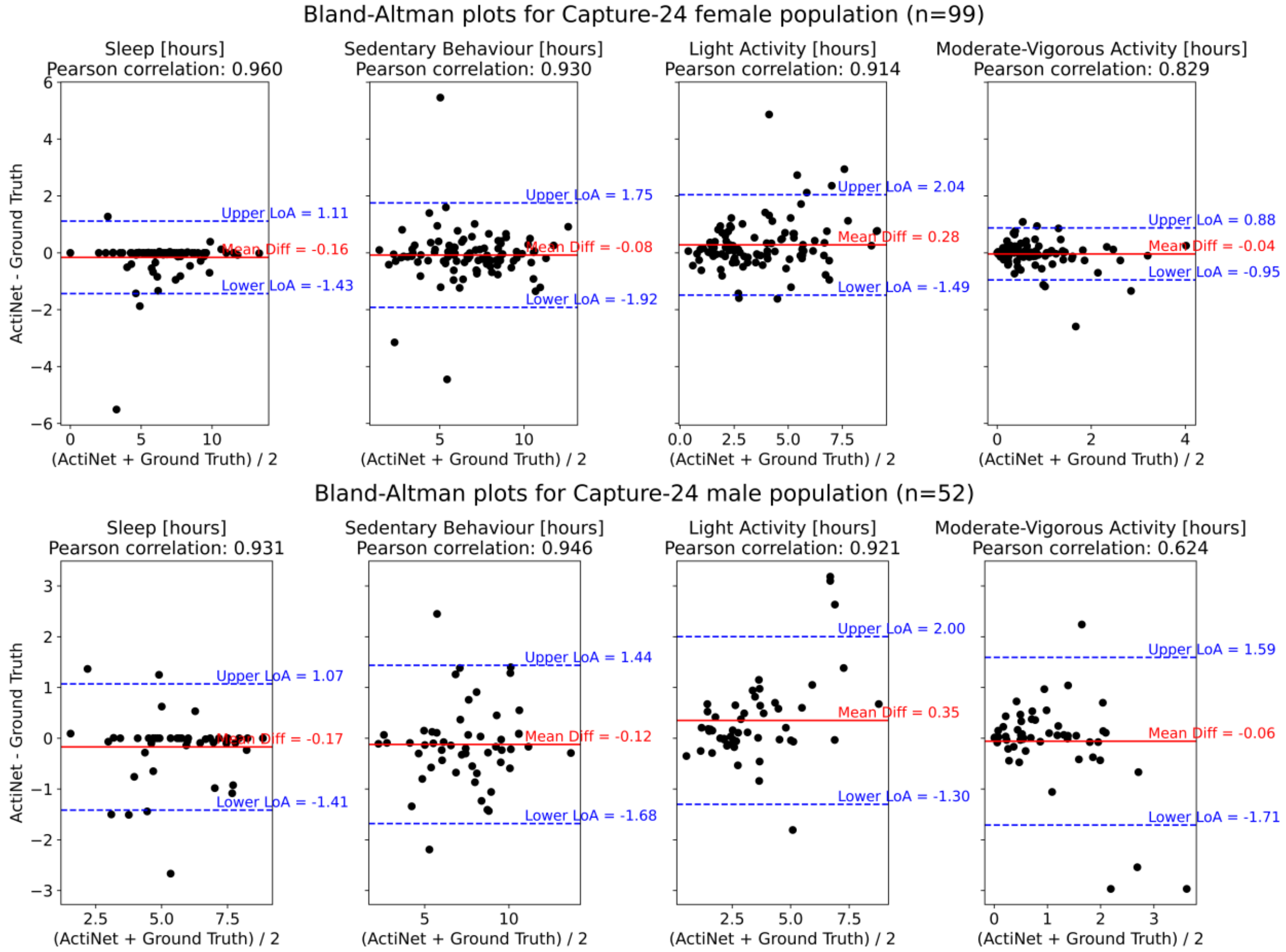

*Supplementary Figure 12: Bland-Altman plots comparing the amount of activity intensity classified by the Baseline and ActiNet models, evaluated on different age bands of the CAPTURE-24 dataset using 5-fold group cross-validation. Each Bland–Altman plot shows the difference between two activity intensity estimates against their mean, with the mean bias (solid red line) and 95% limits of agreement (dashed blue lines) indicating agreement.*

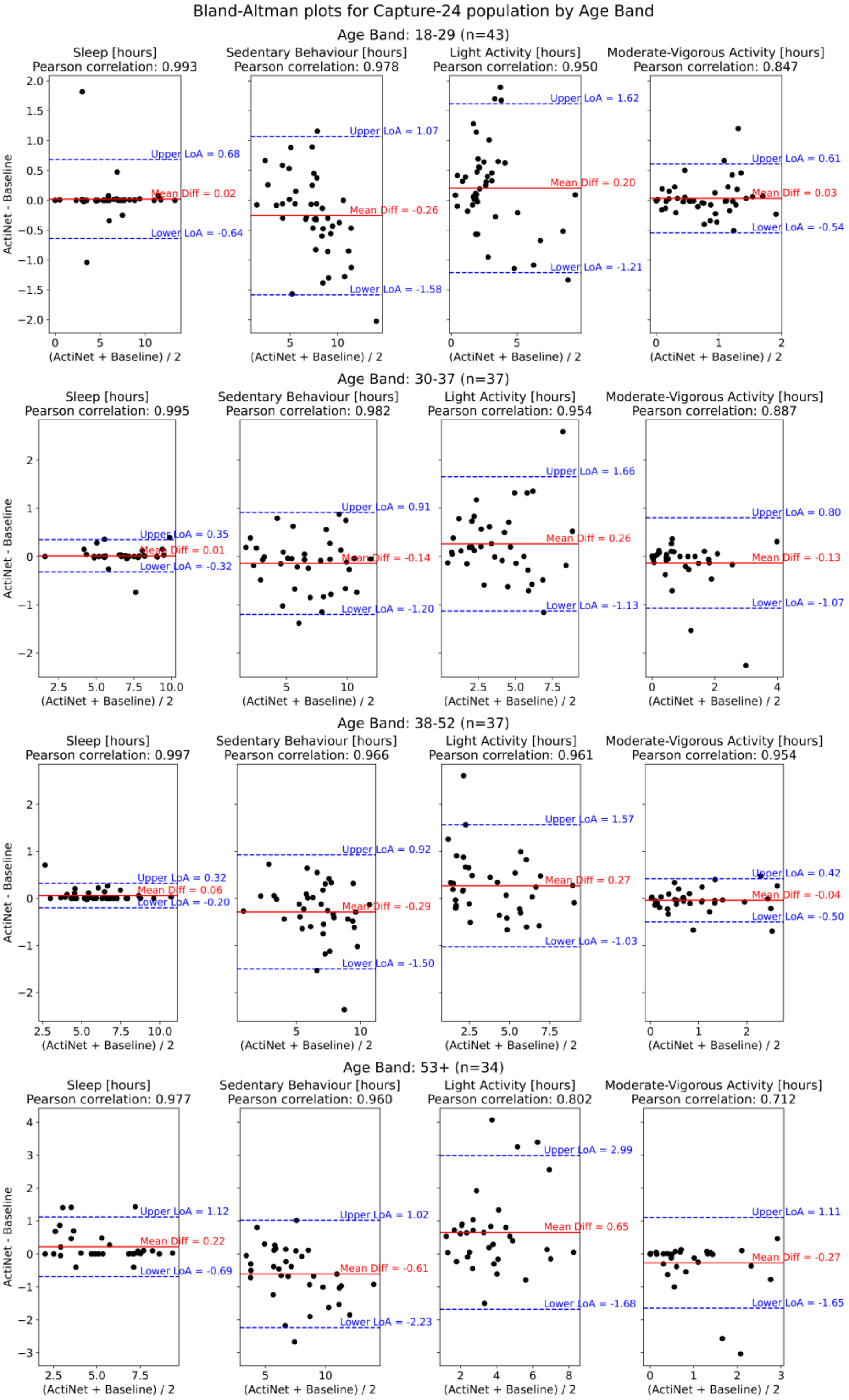

*Supplementary Figure 13: Bland-Altman plots comparing the amount of activity intensity classified by Baseline and ActiNet models, evaluated on difference sexes of the CAPTURE-24 dataset using 5-fold group cross-validation. Each Bland–Altman plot shows the difference between two activity intensity estimates against their mean, with the mean bias (solid red line) and 95% limits of agreement (dashed blue lines) indicating agreement.*

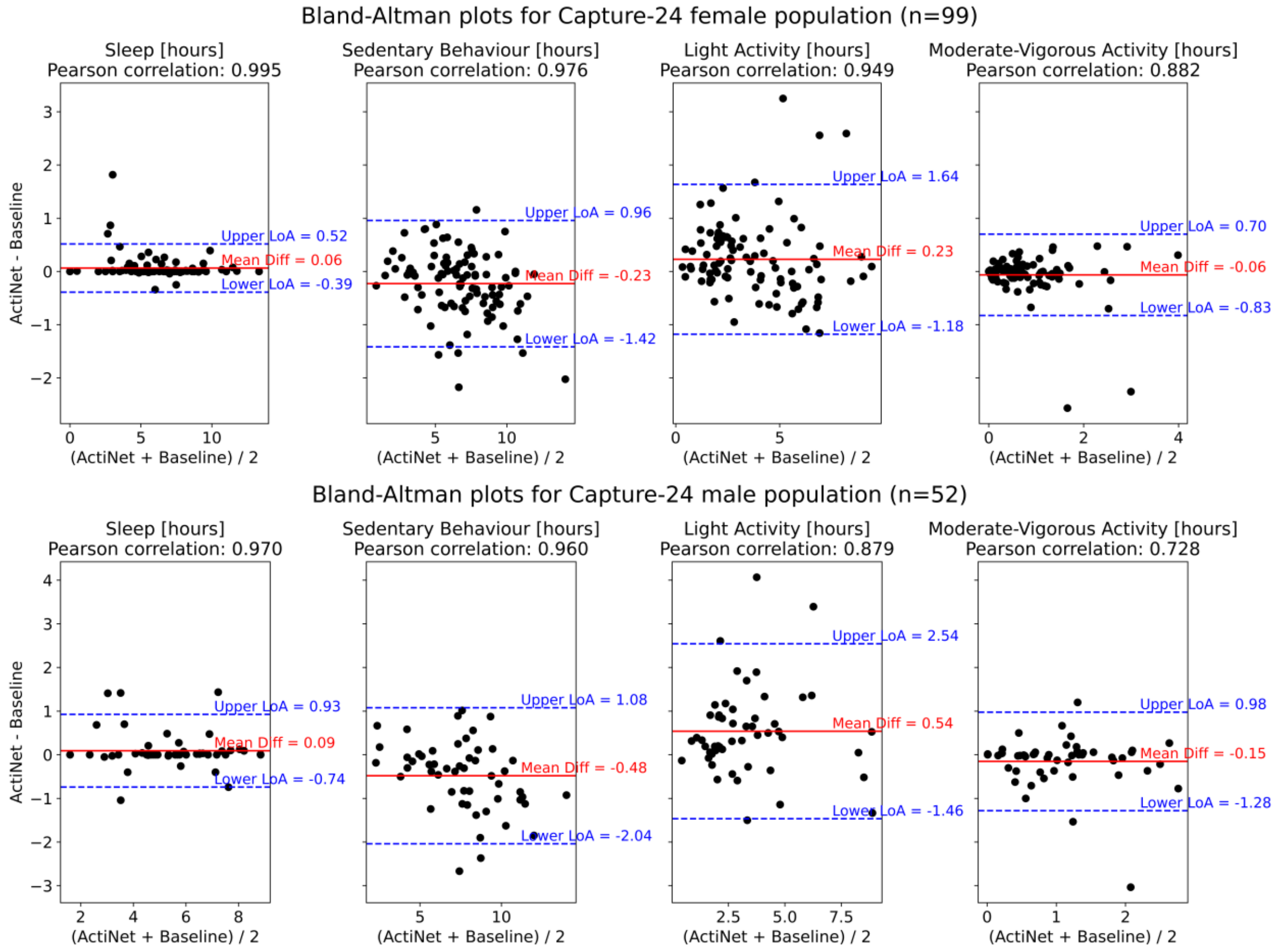